\documentclass[letterpaper]{article}
\usepackage{natbib,alifeconf}

\usepackage{url} 
\usepackage[utf8]{inputenc}
\usepackage{lscape}
\usepackage{color}           % para importar dibujos coloreados

\newcommand{\te}{\vdash}
\newcommand{\lif}{\rightarrow}
\newcommand{\m}{\models}

\newtheorem{theorem}{Theorem}

\newtheorem{definition}[theorem]{Definition.}
\usepackage{graphics} % for pdf, bitmapped graphics files
\usepackage{epsfig} % for postscript graphics files
\usepackage{amsmath} % assumes amsmath package installed
\usepackage{amssymb}  % assumes amsmath package installed

\newcommand{\simon}[2]{{#2}}

\newcommand{\M}{\mathbb M}
\newcommand{\Ob}{\mathbb O}
\newcommand{\A}{\mathbb A}
\newcommand{\I}{\mathbb I}

\newcommand{\ttg}[1]{ #1}

\title{Selecting Attributes for Sport Forecasting using Formal Concept Analysis}
\author{Gonzalo A. Aranda-Corral$^{1}$, Joaquín Borrego-Díaz$^{2}$ \and Juan Galán-Páez$^2$ \\
\mbox{}\\
$^1$Department of Information Technology, Universidad de Huelva, Spain \\
gonzalo.aranda@dti.uhu.es\\
$^2$Department of Computer Science and Artificial Intelligence,
Universidad de Sevilla, Spain\\
jborrego@us.es, juangalan@us.es\\
}

\begin{document}
\maketitle

\begin{abstract}
In order to address complex systems, apply pattern recongnition on their evolution could play an key role to understand their dynamics. Global patterns are required to detect emergent concepts and trends, some of them with qualitative nature. Formal Concept Analysis (FCA) is a theory whose goal is to discover and to extract Knowledge from qualitative data. It provides tools for reasoning with implication basis (and association rules). Implications and association rules are usefull to reasoning  on previously selected attributes, providing a formal foundation for logical reasoning. In this paper we analyse how to apply FCA reasoning to increase confidence in sports betting, by means of detecting temporal regularities from data. It is applied to build a Knowledge Based system for confidence reasoning.
\end{abstract}

\section{Introduction}

Formal Concept Analysis (FCA) \cite{FCA} is a mathematical theory for data
ana\-ly\-sis using formal contexts and concept lattices as key tools.
Domains can be formally modelled according to the extent and the intent of 
each formal concept.
In FCA, the basic data structure is a formal context (with a qualitative
nature) which represents a set of objects and their properties and it is
useful both to detect and to describe regularities and structures of 
concepts. It also provides a sound formalism for reasoning with such
structures, mainly Stem Basis and association rules. 
Therefore, it is interesting to consider its application for reasoning with 
temporal qualitative data in order to discover temporal
trends \cite{Torre}.

In this paper, FCA application scope is the challenge of sports
betting, specifically, the forecasting of soccer league's results.
Forecasting sport results is a fast growing research area,
because of its economic impact in betting markets as well as for its
potential application to problems with similar behaviour (markets)
\cite{Hale}. 
Considering sports betting as a complex system, soccer leagues represent a challenging system with a huge amount of knowledge, available through WWW, and its behaviour is weekly  exhaustive analysed by journalists, betting companies and supporters.
Roughly speaking, three dimensions have been considered for analysing/syn\-the\-si\-zing prediction systems:
1)Those which analyse information on teams (endogenous) versus those which
analyse results (exogenous); 2)Those which exploit quantitative data versus those
which exploit qualitative knowledge, and finally, 3)Statistic-based ones versus
other methods. 
Usually, one can work with hybrid models, and rarely with pure qualitative and
exogenous reasoning systems a\-ppear in literature, although their use is
considered for experiments (for example, frugal methods \cite{frugal} and
based on the recognition heuristic \cite{ecological}) or as part of hybrid
systems (see e.g. \cite{sd4}). 
There are two reasons that may justify this point.

On the one hand, transformation from a large quantitative dataset to a
qua\-li\-tative problem is faced with the selection of an acceptable threshold and the discovery of better relations (see e.g. \cite{bool}). On the
other hand, a qualitative dataset must be accomplished with some amount of
information based on confidence, trust or probability of these data sets.

\begin{figure*}[t]
\centering
\includegraphics[scale=.36]{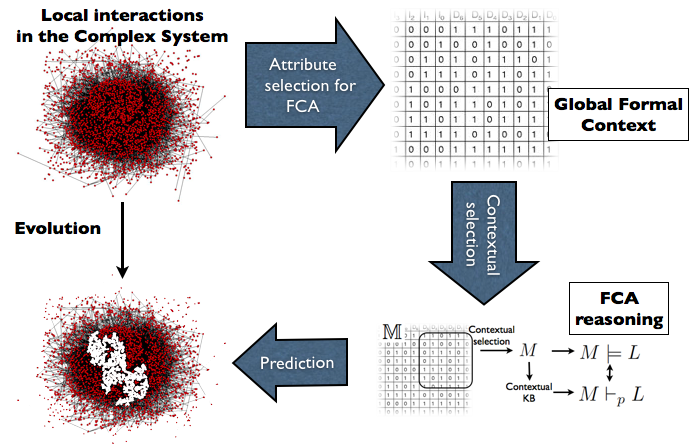}
\caption{FCA based model for prediction of qualitative features of Complex Systems \label{idea}}
\end{figure*}

{\ttg The aim of this paper is to describe all researching work made for selecting and computing attribute sets related to soccer results, {\ttg into a specific framework: FCA, and starting from soccer match results, with no previous analysis of any other specific attributes.} This task is previous to build an Expert System for advising sport betting which could detects some kind of regularities on data. 
Concept lattices, which are computed from attribute values, represent a mathematical structure of relationships among the concepts which are involved in selected sport events to study.
Since this method is bet-oriented, its performance is eva\-lua\-ted within a confidence-based reasoning system. This sistem increases number of hits in soccer matches forecasting, discovering temporal trends by means of data mining and association rules reasoning}. 
The analysis of attributes has been used in \cite{Torre}  to describe a confidence-based  (and contextual) reasoning system for forecasting sports betting. In this paper we analyse the attribute selection problem as a problem of selection of features that shape the behaviour of the complex system that represents professional soccer leagues. 
{\ttg Theoretical framework, on which this model is based on, will be presented at \cite{dali}. 
Due to a really huge amount of information, attribute selection advised by experts is mandatory. 
In fact, the system can be considered as a reasoning model
based on bounded rationality and recognizition heuristics.
and focused on features which were considered as important by human experts.
Therefore, the system aims to forecast results, but it is designed based on bounded rationality models, instead of statistic models (although in the future hybrid models will be considered}.

{\ttg The system is a first prototype from a more general system, which are building to analyse qualitative features of Complex Systems (see Fig. \ref{idea}), using FCA. The idea is to isolate qualitate attributes from (past) local interactions among components of complex system and to apply FCA tools in order to predict properties system's behavior in a near future.}

\section{Background: Formal Concept Analysis}

\simon{According}{According to} R. Wille, FCA \cite{FCA}
mathematizes the philosophical understanding of a concept as a unit 
of thoughts composed of two parts: the extent and the intent. The
extent covers all objects belonging to 
this concept, while the intent comprises of all common attributes 
valid for all the objects under consideration.
It also allows \simon{to  compute}{the computation of}
concept hierarchies from data tables.
  In this section, we succinctly present basic FCA elements (the
fundamental reference is \cite{FCA}). 

\begin{figure*}[t]
\centering
\includegraphics[scale=.45]{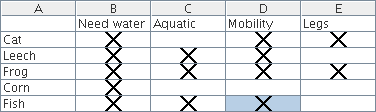} \ \ \includegraphics[scale=0.6]{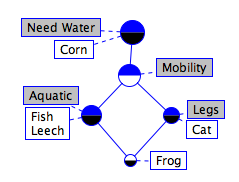} \ \ \includegraphics[scale=.65]{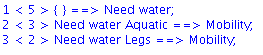}
\caption{Formal context, associated  concept lattice and Stem Basis \label{animal}\label{animales}}
\end{figure*}

A formal context $M = (O, A, I)$ consists of two sets, $O$ (objects) and $A$ (attributes) and a relation
$I \subseteq O \times A$. 
Finite contexts can be represented by a 1-0-table (identifying $I$ with
a Boolean function on $O \times A$). See Fig. \ref{animal} for an example  of formal context about live beings.

The FCA main goal is the computation of the concept lattice associated to the
context. 
Given $X \subseteq  O$ and $Y \subseteq A$ it defines

\begin{center}
$X' :=  \{ a \in A \ | \  o I a \mbox{ for all } o \in X \}$ \\
$Y' :=  \{ o \in O \ | \  o I a \mbox{ for all } a \in Y \} $
\end{center}

A (formal) concept is a pair $(X,Y)$ such that $X'=Y$ and $Y'=X$.
For example, concepts from living beings formal context (Fig. \ref{animales}, left) is depicted in Fig. \ref{animales}, right. 

Using this Fig. \ref{animales}, each node is a concept, and its intension (or extension) can be formed by the set of attributes (or objects) included along the path to the top (or bottom). E.g. The node tagged with the attribute Legs represents to the concept $(\{Legs, Mobility, Need Water\},\{Cat,Frog\})$.

In this paper it works with logical relations on attributes which are valid in the context. 
Logical expressions in FCA are {\em implications between attributes}. An implication is a 
 pair of sets of attributes, written as $Y_1 \to  Y_2$, which is true
with respect to $M = (O, A, I)$  according to the following definition. 
A subset $T \subseteq A$ {\em respects} $Y_1 \to Y_2$  if $Y_1\not\subseteq T$ or $Y_2\subseteq T$.
It says that $Y_1\to Y_2$ holds in $M$ ($M \models Y_1  \to Y_2$) if for all $o\in O$, the set $\{o\}'$
respects $Y_1\to  Y_2$. In that case,  it is said that $Y_1 \to Y_2$ is {\em an implication} of $M$.

\begin{definition} 
Let ${\mathcal L}$ be a set of implications and $L$ be an implication.
\vspace*{-.15cm}
\begin{enumerate}
\item $L$ follows from ${\mathcal L}$  (${\mathcal L} \models L$) if each
  subset of $A$ respecting ${\mathcal L}$  also respects $L$. 
\item ${\mathcal L}$ is complete if every implication of the context follows
  from ${\mathcal L}$. 
\item ${\mathcal L}$ is non-redundant if for each $L\in {\mathcal L}$,
  ${\mathcal L}\setminus \{L\} \not\models L$. 
\item ${\mathcal L}$ is a (implication) basis for $M$ if  ${\mathcal L}$ is complete and non-redundant.
\end{enumerate}

\end{definition}

It can obtain a basis from the {\em pseudo-intents}
\cite{stem} called {\em Stem Basis} (SB):

$${\mathcal L} = \{ Y \to Y'' \  :  \ Y \mbox{ is a pseudointent}\}$$

A SB for the formal context on live beings  is provided  in Fig. \ref{animal} (right). 
It is important to remark that SB is only an example of \simon{basis for a context formal}{a basis for a formal context.} \simon{It}{In} this paper any specific property of the SB \simon{is}{can be} used, and it can be replaced by any implication basis.

It is possible to extend $\models$ \simon{relation}{in relation} to any propositional formula with propositional variables in $A$, by considering each object $o\in \M$ as a valuation $v_o$ on $\A$ defining
$$v_o(A)=1 \Longleftrightarrow (o,A)\in \I$$
Thus $M\models F$ if and only if for any $o\in O$ it holds that $v_o\models F$.

%\subsection{Stem basis and association rule reasoning}

The {\em Armstrong rules} \cite{Armstrong} provides a formal basis for implicational reasoning:
  $$\displaystyle \scriptsize \frac{ }{X \lif X} \ \ \   \frac{X\lif Y}{X\cup Z \lif Y},
\ \ \  \frac{X\lif Y, \ Y\cup Z \lif W}{X\cup Z \lif W}
$$
A set of implications is closed if and only if the set is closed by these
rules \cite{Armstrong}. By defining $\te_A$ as  the proof relation by Armstrong
rules, it holds \simon{that}{that the} implicational bases are $\te_A$-complete:
\begin{theorem}
  Let $\mathcal L$ be \simon{a}{an} implicational basis for $M$, and $L$ an
  implication. Then
    $M\m L$ if and only if ${\mathcal L}\te_A L$
\end{theorem}

In order to work with formal contexts, stem basis and association rules, the
Conexp\footnote{http://sourceforge.net/projects/conexp/} software has been
selected. \simon{It}{It is} used as a library to build the module which provides the
implications (and association rules) to the reasoning module of our system. The reasoning
module is a production system based on which was designed for \cite{hais}. 
Initially it works with SB, and entailment is based on 
the following result:

\begin{theorem} \label{jarr}Let $\mathcal L$ be a basis for $M$
 and $\{ A_1,\dots ,A_n \} \cup Y \subseteq A$.  The following conditions are equivalent: 
  \begin{enumerate}
      \item  $\mathcal S \cup \{A_1,\dots A_n \} \te_p Y $ ($\te_p$ is
        the entailment with the production system). 
      \item $S\te_A A_1,\dots A_n \lif Y$ 
            \item $M \models \{A_1,\dots A_n\} \lif Y$.
  \end{enumerate}
\end{theorem}

\subsection{Association rules for a a formal context}

We can consider a Stem Basis as an {adequate} production system in order to
reason. However, Stem Basis is designed for
entailing  true implications only, without any exceptions into the object set
nor implications with a {low} number of counterexamples in the context.

{Another} more important question arises when it works on predictions. In this case we
are interested {in obtaining} \simon{some methods}{methods} for selecting a result among
all obtained results (even if they are mutually incoherent), and theorem
\ref{jarr} does not provide such a method. Therefore, it is better to consider
association rules (with confidence) instead of true implications and the initial
production system must be revised for working with confidence.

Researching on logical reasoning methods for association rules is
a relatively recent promising research line \cite{balcazar}.
In FCA, association rules are implications between sets of
attributes. Confidence and support are defined as usual. Recall that 
the {\em support}  of $X$, $supp(X)$ of a set of attributes X is defined as
the proportion of objects which satisfy every attribute of $X$, and 
the {\em confidence}  of a association rule is 
$conf(X\to Y) = supp(X \cup Y) / {supp}(X)$.  Confidence can be
interpreted as an estimate of the probability $P(Y|X)$, the probability of an
object {satisfying} every attribute of $Y$  under the condition that it also
satisfies every one of $X$. 
Conexp software provides association rules (and their confidence)
for formal contexts.  

\begin{figure*}[t]
\centering
\includegraphics[scale=.3]{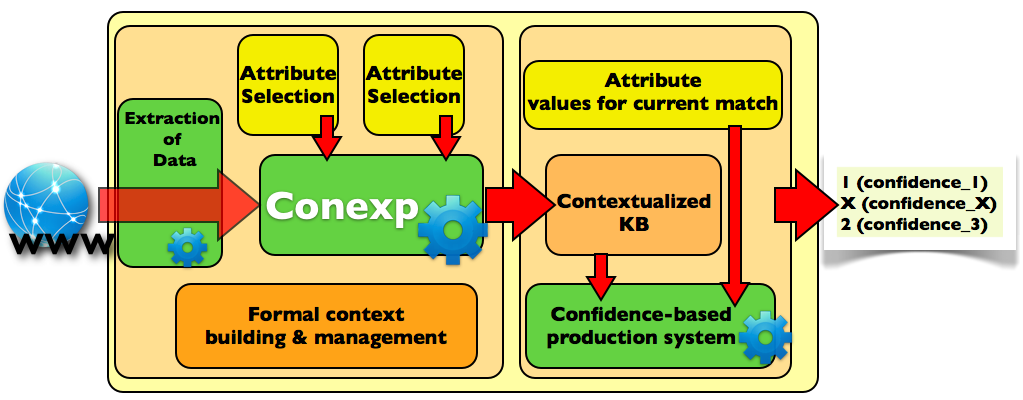}
\caption{Context based reasoning system \label{system}}
\end{figure*}

\section{Reasoning under contextual selection. Logical Foundations}

The model (described in \cite{dali}) is \simon{composed by}{composed of} events (objects) which have a number of properties (attributes). They consitute a {\em universal formal context} $\mathbb M$  (which we call {\em monster context} following the tradition in Model Theory). Thus $\M$ can be considered as the {\em global memory} from which subcontexts are extracted. Once the specific context is considered, it is also possible to consider background knowledge $\Delta$ (in form of propositional logic formulas) which would be combined with the knowledge extracted from formal context (Stem basis or association rules). 

\begin{definition} Let $\M=(\Ob,\A,\I)$ be the monster context, and let $O$ be a set of objects.
\begin{enumerate}
\item A context on $O$ is a context $M=(O_1,A,I)$ where $O\subseteq O_1\subseteq \Ob$
\item A {\bf contextual selection} on $O$ and $M$ is a map $s: O \to {\mathcal P}(O_1)\times {\mathcal P}(A)$
\item A {\bf contextual KB for an object  $o\in O$ w.r.t. a selection $s$ with confidence $\gamma$} is a subset of association rules with confidence  greater  or \simon{equal that}{equal to} $\gamma$  of the formal context associated to $s(o)=(s_1(o),s_2(o))$, that is, to the context 
$M(s(o)):= \displaystyle (s_1(o),s_2(o),I_{\restriction s_1(o)\times s_2(o)})$
(note that when condifence is 1 the contextual KB is a implicational basis).
\end{enumerate}
\end{definition}

Contextual KBs is useful for entailing attributes on an object. 
%A kind of formal contexts are the {\em temporal context on a set of objects} \cite{Torre} (see Sect. 4 bellow).
The reasoning model on $\M$ is argumentative, where \simon{argument}{the argument} is based on KBs extracted from subcontexts \cite{dali}:

\begin{definition}
Let $L$ be an implication and $\Delta$ a background knowledge. It is said that $L$ is a possible consequence of $\M$ under $\Delta$, $\M \models^{\Delta}_{\exists} L$, if there exists $M$ a nonempty subcontext of $\M$ such that $M\models \Delta \cup \{ L \}$. 
\end{definition}

Note that by theorem \ref{jarr}, when $\Delta$ is a set of implications,  it holds that $\models_{\exists}$ is equivalent to $\te_{\exists}$ which is defined by:  $\M \te_{\exists} L$ if there exists $M\models \Delta$ a subcontext of $\M$ such that $S\te_p L$ (where $S$ is a stem basis for $M$).

\subsection{The role of attribute selection for formal contexts}

%Los atributos son fundamentales para la contextual selection y  para construir los contextos. De los contextos se
%saca el conjunto de association rules which are used by the production system. EXPLICAR ESTO.
%
%Los Atributos constituyen pues una de las partes más importantes y sensibles del sistema. Digo sensible porque de lo bien
%que estos atributos representen el comportamiento de los equipos dependerá la calidad de los resultados obtenidos.

{\ttg Attributes are essentials in the contextual selection to build good formal contexts. Association rules are extracted from
the contexts and those are used by the production system. By means of these association rules and some initial facts based on the
match we want to forecast the production system infers the confidence (probability) for each one of the three possible results of a match,
home team wins, draw or away team wins.
%¿sensitive es la traduccion correcta de sensible?
Thus attributes constitute one of the most important and sensitive parts of the system. They are sensitive because on
how they represent the behavior of the teams will depend the accuracy of the inferred results.}

\begin{figure*}[t]
\centering
\includegraphics[scale=.36]{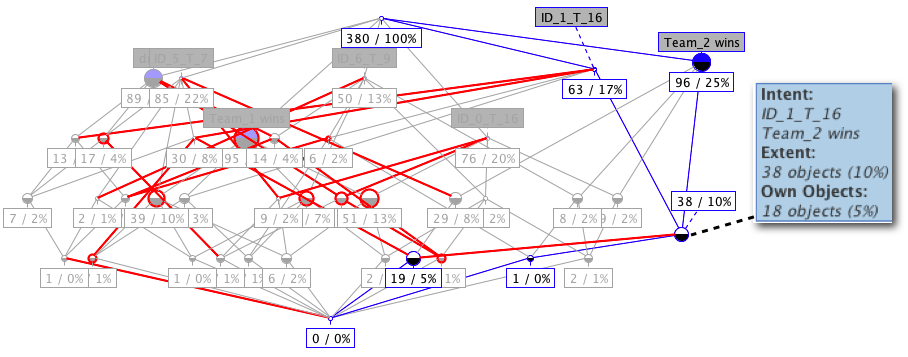}
\caption{Concept Lattice for  the match {\em Málaga-Sevilla} (week 31, season 2009-10) \label{reticulo}}
\end{figure*}

\section{Confidence-based reasoning system}

The reasoning system works on facts of the type $(a,c)$, where $a$ is an attribute and $c$ is the estimated probability of the trueness of $a$, which we also call confidence (by similarity with the same term for association rules).  See \cite{Torre} for a more detailed description of the reasoning system.

The system has a module {for a} confidence-based reasoning system (Fig. \ref{system}). Its entries for a match $Team_1$ - $Team_2$ are: the contextual Knowledge basis for a threshold given as rule set and attribute values for the current match (except 1,X,2) as facts, all of them with a confidence (whose value depends on the reasoning mode, see {below}).  The production system is executed and the output is a triple $<(1,c_1),(X,c_x),(2,c_2)>$ of attribute, confidence for this match. The attribute with greater confidence is selected as the prediction. Production system execution is standard, with several modes for confidence computing of results based in uncertain reasoning in Expert Systems \cite{giar}. 
Any attribute/fact $a$ is initialized with confidence
$$conf(a):=\displaystyle\frac{|\{o \ : \  oIa \}|+1}{|O|+1}$$

%%%PANTALLAZO%%%

%Fig. \ref{pantallazo} shows forecasting for week 21 of {the Spanish premier league} (2009-10).
%%It only provides two automatic selection modes. The default mode (mode 1) $0,5$ and the more interesting (mode 2)

%
%\begin{figure*}[t]
%\centering
%\includegraphics[scale=.5]{pantallazo}
%\caption{Forecasting results screenshot
%%Correct predictions on 2009-10 league 
%\label{pantallazo}}
%\end{figure*}

%%%%%%%%%%%%%%%%%%%%%%%%%%%%%%%%%%%%%%%%%%%%%%%%%%%

\section{Attributes and formal contexts for soccer league}

For both selecting data and building contexts, some assumptions on forecasting in soccer league matches {have been} considered. Reconsiderations of such {decisions} can be easily computed in the system.
First, we consider that the regularity of team's behaviour only depends on the contextual selection that {has} been considered. 
This contextual selection is obtained by taking matches from the last $X$ weeks backwards,
starting from the week just before the one we want to forecast.%, where $X$ is the size of the seasson.
Second, since FCA methods are used {to} discover regularity features, thus it does not {consider} forecasting exceptions (unexpected results). Therefore, {the model can be considered as a starting point for betting expert who would adjust attributes, in order to more personalised criteria.}

These attributes have to be computed and used to entail the forecasting. This analysis is assisted by Conexp.
ConExp software is used to compute and analyze the concept laticces associated to the temporal contexts. In order to select most interesting attributes for the system, starting from an initial configuration, user can compute the associated concept lattice and check it. In this way, attributes goodness (and thresholds) can be evaluated to reconsider current attribute selection. 
For example, in Fig. \ref{reticulo}, the concept lattice associated to contextual selection for {\em Málaga-Sevilla} match is shown. 
This contextual selection is obtained from a given attribute selection and last  38 weeks matches before. 
In this concept lattice, the attribute $ID\_1\_T\_16$ is defined by:
'the budget of $team_2$ is greater than $\gamma_1$ times the budget of $team_1$', where $\gamma_1$ is the threshold the expert must estimate.
In the concept lattice we can observe that the biggest concept containing the attributes $team_2\_wins$ and $ID\_1\_T\_16$ covers the about the 10\% of the
objects owned by the first attribute, therefore it is suggested to use the second attribute for reasoning with association rules to get a prediction.

{The system computes the  value of an amount of attributes on objects}. Experimentally {a boolean} combination of attributes is possible. {Once the temporal context has been computed}, the system can build contextual selections by selecting the match and the attribute set. %(see fig. \ref{muestra})
The selection of attributes was made by considering four kinds of factors: those related with the classification, {the} history of teams' matches in the recent past, results of {direct} matches and other non related {results,} as for example the difference between team budgets. {Seventeen relevant attributes were selected}.%, some of them parameter-dependent. 
The attribute set has three {special attributes}, $Team_1$ wins  (1), $Team_2$ wins (2) and {draws} (X). 

{\ttg With respect to data source, they are automatically extracted
from RSSSF Archive\footnote{http://www.rsssf.com}. 
Objects are matches and attributes are a list of features, including temporal stamp (week, year). Data was collected for the past four years. Actually the size of the context is about 300 objects and 18 attributes (although several of them are parametrized, see section bellow). Thus, $|I|$ is about 5,100 pairs.}

\section{Attribute selection}
{\ttg
We have chosen a small set of attributes with many possibilities through a few customizable parameters. When these parameters
are having set up with proper values, the set of attributes will represent team's logical behavior.
%represent<->simulate?

%He elegido un conjunto de atributos pequeño, pero con gran cantidad de posibilidades, que si son configurados de forma
%adecuada pueden ajustarse en cierta medida al comportamiento de los equipos. 

Recall that formal concept analysis works with qualitative attributes and all teams information which we work with are quantitative data. Thus it is necessary to convert quantitative attributes into qualitative ones. This task is left to users by choosing a proper threshold to each attribute.

%Recordemos que la técnica de Análisis Formal de Contextos trabaja con Atributos Booleanos, mientras que
%las propiedades de un equipo las medimos con valores discretos. Cuando tenemos un Atributo que representa una propiedad
%de un partido, este atributo irá acompañado de un Límite o threshold, a partir del cual obtendremos el valor booleano
%del atributo, según si el valor discreto sobrepasa el threshold o no.

%revisar
Before choosing the set of base attributes , we have carried out a analysis on information about soccer results . The aim have been to discover which factors are more influential in teams behavior and which ones are less influential. 
First of all, we have collected any interesting factor found, and after analyzing each one, individually, we have chosen most suitable ones. 
Examined factors can be classified in four different categories (see Table \ref{others}): those related to season's classification, those related to previous team's results, those related to historical direct matches and any other factors.
It is worth to note that to increase possibilities of the attribute set, and considering the Boolean nature of  formal context attributes, we have added the option to create new ones by means of logical combinations of these attributes.}

%Para obtener el conjunto de atributos se ha realizado un estudio previo sobre la información del mundo del futbol que existe
%a nuestra disposición. Este estudio previo incluye el análisis del comportamiento de los equipos para obtener qué propiedades
%reflejan este comportamiento y cuáles no. A priori se ha tenido en cuenta cualquier factor posible y luego se han ido seleccionando según su eficacia.
%A continuación se listan los factores analizados detallando el grado de correlación que hay entre
%la información que aportan los factores y el resultado de los partidos y cuales se han tomado y cuáles no. 

%{\tt It is worth to note that Para hacer todavía más completo el conjunto de atributos y teniendo en cuenta que los Atributos son booleanos, le he añadido
%la posibilidad de crear Atributos compuestos a partir de otros atributos, realizando combinaciones lógicas entre varios de estos}

%{\tt Los factores que se han considerado para la selección y construcción de atributos son de cinco tipos (see Table \ref{others}):
%relacionados con la clasificación,  relacionados con las trayectorias de resultados, Those relacionados con los enfrentamientos directos, and others.}

%\subsection{Attribute set}

{\ttg According to considered factors, 
the system computes a base set of 18 attributes, which are customizable by some parameters. This will
let us to obtain a diverse set of attributes. In Table \ref{att} attributes are specified}. {\ttg Four parameters are used}:

\begin{itemize}
\item {\ttg Threshold: Parameter to be used to translate quantitative attribute values into qualitative ones.}
%es valor límite a partir del cual se considera que el objeto satisface el Atributo, lo que nos permite convertirlos en valores booleanos. Toma valores enteros a partir de 1.
\item {\ttg Team: Recall that in the formal context considered, objects are matches but attributes belongs to team properties.
This parameter will set the team from object (match) on which attribute will be considered. It has two possible
values: \{HOME, AWAY\}. Thus, usually, we will have twice each attribute at context, once for home team and once for away.}
%Recordemos que en nuestro contexto los objetos son partidos, y sin embargo las propiedades siempre se miden sobre un equipo.
%Este parámetro sirve para indicar al sistema sobre cuál de los dos Teams del objeto se aplica el Atributo.
%Conjunto de valores que puede tomar: \{LOCAL, VISITANTE\}.

%¿set<->stablish?
\item {\ttg Number of Matches: sets the number of past matches to be considered when some attributes are computed, e.g. the ones associated to previous team results.}
%En los Atributos en los que se miden propiedades relacionadas con un cierto número de partidos pasados, este parámetro
%sirve para configurar el número de partidos pasados que la aplicación debe analizar para calcular el valor del Atributo. Toma
%valores Enteros a partir de 1.
\item {\ttg Kind of matches: sets past matches type to be taken into account to compute some attributes, considering home/away team's condition at matches. Three possible values: \{MATCHES AS HOME TEAM, MATCHES AS AWAY TEAM, ALL MATCHES\} }
%En todos los atributos (menos en los relacionados con el presupuesto) podemos indicar qué tipo de partidos
%debe tener en cuenta la aplicación al calcular los Atributos. En algunos casos puede que nos interese por ejemplo mirar la
%trayectoria del equipo en las últimas semanas teniendo
%en cuenta solo los partidos en los que juega como local. El conjunto de valores que puede
%toma este parámetro es  \{TODOS, COMO LOCAL, COMO VIyesTANTE\}.
\end{itemize} 

{\ttg 
%¿versatile?
With these parameters, and the possibility to compound attributes, it is possible to build a detailed attributes set.
Note that experiments show that simplest and most logical attributes give a good team behavior representation.
Although we consider that a versatile attributes set, as above described, was necessary because of a huge number of factors can 
determine the result of a soccer match.
%¿El paqrrafo siguiente? ¿quitar/modificar?.
Task of customizing the attribute set is left to users, and it is the most important one in forecasting process. Thus,
a basic soccer knowledge should be required. The goodness of customization will determine system results.
}

%\begin{landscape}
\begin{table*}[t]
 \begin{tabular}{|p{7.45cm}|c|c|}\hline
\bf Factors & \bf Correlation Degree & \bf Used? \\ \hline
\bf Associated to the classification in the league & \bf  & \\ \hline
Team in the first classification level & medium/high & yes \\ \hline
Team in the last classification level & medium/high & yes
\\ \hline 
%Diferencia entre las posiciones  de los dos equipos en la clasificación
Difference between team's classifications & medium/high & yes
\\ \hline Team was in a different league last year  & medium & no
\\ \hline Team socred a important number of goals (in the last matches)& medium/low & no\\ \hline
%\end{tabular}
%\caption{Factors associated to the classification considered for building attributes \label{clas}}
%\end{table*}

%
%\begin{table*}
%\centering \begin{tabular}{|c|c|c|}\hline
\bf Associated to previous results of the team &   &  \\ \hline
%Racha de victorias consecutivas que arrastra el equipo.
Number of consecutive won matches.
& high
& yes
\\ \hline Number of consecutive lost matches.
%Racha de derrotas consecutivas que arrastra el equipo.
& high
& yes
\\ \hline Number of consecutive draws.
%Racha de Drawns consecutivos que arrastra el equipo.
& medium
& yes
\\ \hline Number of non consecutive won matches in previous weeks.
%Victorias cosechadas por el equipo en los últimos partidos.
& high
& yes
\\ \hline Number of non consecutive lost matches in previous weeks.
%Derrotas cosechadas por el equipo en los últimos partidos.
& high
& yes
\\ \hline Number of non consecutive draws in previous weeks.
%Empates cosechados por el equipo en los últimos partidos.
& medium/high
& yes
\\ \hline Points collected in previous weeks.
%Puntos obtenidos por el equipo en los últimos partidos.
& medium/high
& yes\\ \hline
%\end{tabular}
%\caption{Factors associated to the previous results of the team \label{clas}}
%\end{table*}

%\begin{table*}
%\centering \begin{tabular}{|c|c|c|}\hline
\bf Factors related with directed matches (incluidas previous years)&  & \\ \hline
Number of wins in previous directed matchs & medium/high & yes
\\ \hline number of losts in previous directed matchs & medium/high & yes
\\ \hline number of draws in previous directed matchs & medium/high & yes \\ \hline
%\end{tabular}
%\caption{Factores relacionados con los enfrentamientos directos anteriores (incluidas ligas previas)\label{direc}}
%\end{table*}

%\begin{table*}
%\centering \begin{tabular}{|c|c|c|}\hline
\bf Other Factors &  &  \\ \hline
Number of red cards collected by the team's players.
%Tarjetas rojas que reciben los jugadores del equipo.
& low
& no
\\ \hline Wheather the day and the city where the match took place & medium  & no (hard to parametrize)\\
%Clima el día del partido
\hline Motivation because of the fans support when playing as home team.
%Efecto jugar en casa, afición animando y asistencia media al estadio
& high
& no (hard to parametrize, subjective)
\\ \hline Team hires a new coach.
%Se produce un cambio de entrenador en el equipo.
& high
& no (only useful when new coach hired)
\\ \hline Some players of the team are selected for their National Team.
%El equipo tiene jugadores que juegan en selecciones nacionales
& medium/Low
& no (relevant for some nationalities)\\
 \hline Difference between team's budgets.
%Diferencia de presupuesto entre los dos equipos que se enfrentan.
& high
& yes
\\ \hline One or more important team's players are injured.
%Lesiones sufridas por los jugadores más relevantes de un equipo.
& medium
& no (hard to automatically collect the data)
\\ \hline Cups collected in the lasts years.
%Títulos cosechados por los equipos a lo largo de la historia.
& low
& no (only for a few of teams) \\ \hline
\end{tabular}
\caption{Factors considered for selecting/building attributes \label{others}}
\end{table*}

\section{Computing problems}

%La dinámica propia de la competición hace necesaria reconsiderar en una serie de situaciones los cálculos de los valores de
%los atributos a partir de la base de datos. En esta sección se describen los principales problemas  de este tipo que han
%surgido. Roughly speaking, the main problems concerns to the initial matchs in soccer league.
{\ttg
The way of competition  causes to take into account some special situations for computing attributes values. In this
section we describe the main problems emerged and how they were fixed. Roughly speaking, these main problems concerns to
initial matches in season.}

%Jornada 0 de Liga
\subsection{Beginning of a new season: week 0}
{\ttg
This problem is not hard, but as many others unavoidable, and a solution becomes essential. It happens when computing an attribute value related to league standings to forecast first week of a season.
As any previous week has been played yet, there is not way to build a standing table.

When teams in current season remain in the same league as last, a trivial solution is to take into account positions and matches in last weeks of previous season. If the team played in a higher division than last season, it will be at the first position in the standing. Otherwise, if the team played in a lower division, it will be considered at last position.}

%Este problema es relativamente poco importante. Ocurre cuando queremos calcular atributos para consultar un partido de
%la Jornada 1. Los Atributos conflictivos son los relativos a la clasificación actual, y en ese momento no existe. 
 
%La solución elegida consiste en 
%tomar la clasificación de la última jornada
%de la temporada anterior para esa división. Con respecto a los equipos que han ascendido o descendido a esa división no están incluidos en esa clasificación, la solución  elegida
%consiste en que ocupen todos ellos la misma posición,
%en el caso de los descendidos será la primera posición, y en el caso de los ascendidos la última.

%Primeras Jornadas de Liga y Falta de Matchs para los cálculos
%ESTAS DOS SECCIONES SE HAN UNIFICADO.
\subsection{Missing matches in attribute computation}
{\ttg 
%el it happens thet no suena nada bien.
Other problem, closely related to previous one, is when not enough previous matches are available to compute an attribute. 
Solution pass through taking lasts matches of last season as if they were in a continuous temporal line. This
is not so simple, because of some teams were not playing at same division last season. 
Indeed, when playing in a lower or higher division,  difficulty of division changes and matches cannot be compared into the same way. 
Therefore, we need to handle the situation of a team playing in a different division from current season division.

Other troubled situation where there are not enough matches for attribute computation is to compute results for directed matches between two teams because of there is only a few of such matches in the data source.

%le estamos diciendo que no cumple el atributo cuando en realidad, no tenemos informacion para decir si lo cumple o no lo cumple
For these two related situations we offer two solutions. First is to compute attribute with a null value,
but in this way we are giving a fake information to the system. We are setting that attribute is not true but, in fact, we have not information enought to determine it, so a better approach is required. Chosen solution  is based on adjusting attribute's threshold. The value of this threshold is decreased proportionally to relation between
number of required matches and number of available matches. Threshold $\gamma$ is revised by

$$\gamma_{new} = \gamma_{old} \cdot \frac{\mbox{number of match results available}}{\mbox{number of match results needed}}$$

When number of required matches is too high and number of available matches is low, it looks like we are giving fake
information to system again, but our experience shows that collateral effects of this approach are worthless compared to compute attributes with a null value.}

\section{Attribute selection vs expert system behaviour}
{\ttg 
%ESPECIFICAR ESTO DE CONDICIONES NORMALES
In general terms, current base attribute set behavior forecast the most possible results of a match is quite
good, in regular conditions. Even so, some experiments, in order to study attribute's behavior, have been developed.
%after  following observations about  in expert system have been taken.} {\Large * * * NI PUTA IDEA!!! * * *}

%En general el comportamiento del conjunto de atributos actual a la hora de aproximar los resultados de un partido es bastante
%bueno en condiciones "normales". Aun así se han realizado las siguientes observaciones en torno al comportamiento de
%los Atributos en el Sistema Experto.

%%Esto son en parte resultados de la experimentación.
\subsection{Strict attributes}\label{conjuntoEstricto}
{\ttg 
An attribute is \textit{strict} when only a few objects can satisfy it, because of its threshold is too high. By working with
sets of strict attributes, we can assure that they estimate the teams behavior better than other sets. 
Thus, with strict attributes, we will have very reliable estimates, but just only for very few matches, and non for most of others. 
In the other hand, using less strict attributes, system will produce less reliable estimations but for a big scope. 
So it is essential to find a balance between these two opposite situations: reliability of attribute set against number of matches without information.
A good solution could be to build and use different attribute sets, ones more strict and others less. Thus, less strict attribute sets
will be used when strict ones fail doing an estimation.}

\begin{table*}[t]
 \begin{tabular}{|p{8cm}|p{8.5cm}|}\hline
\bf Attribute & \bf Configurable parameters  \\ \hline
  1) Number of non consecutive won matches in previous weeks $>$ threshold
& $<$threshold$>$ $<$Team$>$ $<$Number of Matchs$>$ $<$Matchs$>$ 
\\ \hline 2) Number of non consecutive lost matches in previous weeks $>$ threshold
& $<$threshold$>$ $<$Team$>$ $<$Number of Matchs$>$ $<$Matchs$>$ 
\\ \hline 3) Number of non consecutive draws in previous weeks $>$ threshold
& $<$threshold$>$ $<$Team$>$ $<$Number of Matchs$>$ $<$Matchs$>$  
\\ \hline 4) Points collected in previous matches$>$ threshold
& $<$threshold$>$ $<$Team$>$ $<$Number of Matchs$>$ $<$Matchs$>$ 
\\ \hline 5) Position in the classification based on previous matches$>$ threshold
%Clasificación según últimos partidos: posición
& $<$threshold$>$ $<$Team$>$ $<$Number of Matchs$>$ $<$Matchs$>$ 
\\ \hline 6) Number of positions over the opponent in the classification based on previous matches$>$ threshold
%Clasificación según últimos partidos: puestos por encima del Rival
& $<$threshold$>$ $<$Team$>$ $<$Number of Matchs$>$ $<$Matchs$>$ 
\\ \hline 7) Number of positions under the opponent in the classification based on previous matches$>$ threshold
& $<$threshold$>$ $<$Team$>$ $<$Number of Matchs$>$ $<$Matchs$>$  
\\ \hline 8) Number of wins in previous directed matchs (included previos leagues) $>$ threshold
& $<$threshold$>$ $<$Team$>$ $<$Number of Matchs$>$ $<$Matchs$>$  
\\ \hline 9) Number of losts in previous directed matchs (included previos leagues) $>$ threshold
& $<$threshold$>$ $<$Team$>$ $<$Number of Matchs$>$ $<$Matchs$>$ 
\\ \hline 10) Number of drawns in previous directed matchs (included previos leagues) $>$ threshold
& $<$threshold$>$ $<$Number of Matchs$>$ $<$Matchs$>$ 
%Clasificación hasta jornada Actual: posición
\\ \hline 11) Position in the classification $>$ threshold
%Clasificación hasta jornada Actual: puestos encima de Rival
& $<$threshold$>$ $<$Team$>$ $<$Matchs$>$
\\ \hline 12) Number of positions over the opponent in the classification$>$ threshold
& $<$threshold$>$ $<$Team$>$ $<$Matchs$>$
\\ \hline 13) Number of positions under the opponent in the classification$>$ threshold
& $<$threshold$>$ $<$Team$>$ $<$Matchs$>$ 
%Victorias consecutivas hasta la fecha
\\ \hline 14) Number of consecutive won matches$>$ threshold
& $<$threshold$>$ $<$Team$>$ $<$Matchs$>$
%Derrotas consecutivas hasta la fecha
\\ \hline 15) Number of consecutive lost matches$>$ threshold
& $<$threshold$>$ $<$Team$>$ $<$Matchs$>$
%Empates consecutivos hasta la fecha
\\ \hline 16) Number of consecutive draws$>$ threshold
& $<$threshold$>$ $<$Team$>$ $<$Matchs$>$
\\ \hline 17) Team's budget Y times bigger than opponent's budget (Y $>$ threshold)
%Presupuesto Y veces mayor que el del rival
& $<$threshold$>$ $<$Team$>$
\\ \hline 18) Team's budget Y times smaller than opponent's budget (Y $>$ threshold)
%Presupuesto Y veces menor que el del rival
& $<$threshold$>$ $<$Team$>$ \\ \hline
\end{tabular}
\caption{Attributes and parameters\label{att}}
\end{table*}
%\end{landscape} 

\section{Trends towards the victory of the home team}
{\ttg
It is a fact that, in soccer, it is more probable a victory from home team than away team. To deal with this, we offer two
different approaches. First, modelling the teams behavior and second computing confidence values.
%ESTA FRASE NO SE COMO PONERLA.
For modeling  teams behavior (attribute set customization) it is a good practice to use attributes with low exhaustive thresholds for home team and more exigent threshold for attributes related to away team. Therefore, it will be easier for home team
to satisfy an attribute than away team. It is possible to imitate this trend based on this approach.

Around 50\% of played matches finish with victory of home team. This means that the attribute value, corresponding to matches result, will be 'home team victory' around 50\% of objects from formal context. As consequence of the former, many rules from the inferred association rules will contain the attribute 'result = home team victory' within their conclusions. 
Thus when forecasting a match the system will infer, in most of cases, 'home team victory' as consequence of overestimation confidence value for this result. 
It is possible to avoid  this effect easily, just applying a decreasing (reduction) factor over confidence for 'home team victory'. It is estimated by means of experiments. }

\begin{figure}[t]
\centering
\includegraphics[scale=.57]{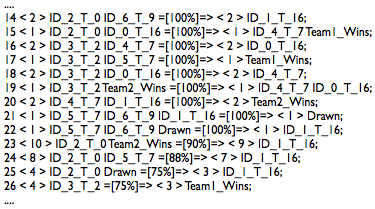}
\caption{KB fragment from Fig. \ref{reticulo} \label{KBMS}}
\end{figure}

\section{Results}

Following the process described above, \simon{it ran an experiment}{an experiment was run}  {for the Spanish premier soccer league} from
2009-10. Attributes \simon{are}{were} selected according the experience of an expert, and contextual KB is computed (in Fig. \ref{KBMS} a KB fragment for {\em Málaga-Sevilla} match is shown).  From this
selection $\te_{\exists}$ is computed for each match in each week. 
%\begin{figure*}[t]
%\centering
%\includegraphics[scale=.45]{grafica}
%\caption{Correct predictions on 2009-10 season of first league \label{grafica}}
%\end{figure*}

%HE PUESTO LAS DOS GRAFICAS, PORQUE NO SE CUAL ES MAS INTERESANTE, CUALQUITARIAIS? LA NUEVA EL PROBLEMA QUE TIENE ES QUE SON SOLO 17 JORNADAS.

\begin{figure*}[t]
\centering
\includegraphics[scale=1]{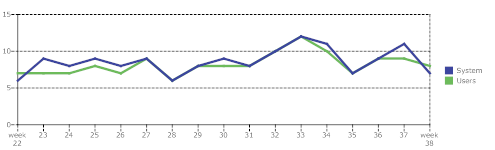}
\caption{Correct predictions on the last 17 weeks of the season 2010-11 \label{graficoExperimento}}
\end{figure*}

\noindent {\bf 2009-10 season:} Experiments with the system show forecasts {of about} 58.16\% by a contextual selection based {on the } {previous 38} matches of each
team. 
%(Fig. \ref{grafica}).
Such \simon{an}{a} percentage of hits for a qualitative reasoning system may be considered as an acceptable
result comparable with expectable results {of} experts \cite{frugal,frugal1}. Experiments with other contextual selections
shows an \simon{increasing of} {increase in} the number of hits \simon{about}{by about} 7\% in the second half of the season. The reason is
that data from the first half provides more recent information on teams and past matches. 

\noindent {\bf 2010-11 season:} {\ttg According to the idea commented above, we have evaluated the system in the second half of 2010-11 soccer season.
A way to evaluate how good is this forecasting sistem is comparing number of successes in our pool with the most popular betting selections. This popular selections are collected from the most voted results for each match, published at state agency web that controls soccer pools.
In Fig. \ref{graficoExperimento}  both results are compared. Our hits are in blue and popular ones in green and last seventeen weeks from 2010-11 season are represented. Note that Spanish soccer pools are over 15 matches.
}
 
% {\ttg 
%temporada = session! 
 
%en la temporada 2010-2011, the refinement in the selection of attributes ha dado lugar a... sobre seis apuestas que son elegidas para...}

%en la jornada 5 se obtiene una de 10 (0 eur.), en la jornada 21, 12 aciertos (569 eur.), en la jornada 31, 10 acierto (14,42 eur.).  El total de
%premios sería 583,42 eur. (501 de beneficio neto).

%\vspace*{-.15cm}

\section{Conclusions and Future Work}

\begin{figure*}[t]
\centering
\includegraphics[scale=.28]{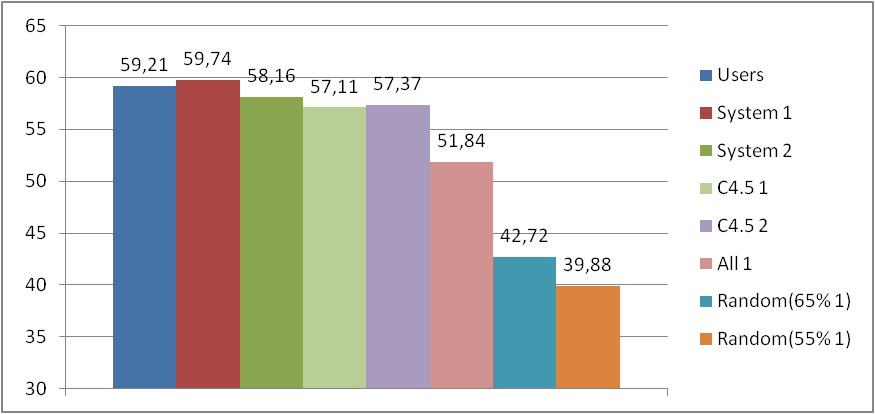}
\caption{Comparative of correct predictions on the whole season 2010-11. Percentages \label{comparativaC45}}
\end{figure*}
%REVISAR CAPTION!!!

{\ttg 
The challenge to detect emergent concepts for reasoning about complex systems represents an exciting researching field. Concepts with qualitative nature are extracted from data only considering partial features of complex system dynamics,  a partial understanding of system. 
In this paper, FCA is applied to this aim with a specific application}.
The selection attribute problem based on FCA-based reasoning system for sport forecasting is analysed. In fact, the reasoning system is a computational logic model for bounded rationality. The model \simon{concerns}{is concerned} with association rule reasoning and it does not use -in its current form- more sophisticated probability tools (as for example \cite{sd4}). As is stated in \cite{prob}, the theory
of probabilistic mental models assumes that inferences about
unknown states of the world are based on probability cues
\cite{Bru}. It can say that confidence of association rules extracted from subcontexts play the role of probability cues.

Any statistical approaches have been taken into account, because of it was not the aim of this paper. Although a comparative study of our system against C4.5 classifier has been done. 
For this, two different attribute selections have been considered and used for both, C4.5 classifier and our system. 
The experiment is to forecast all matches (380) in season 2010-11. 
In order to stimate each match result, considering $N$ (weeks) as timestamp, previous matches are used to build contextual selection (or trainining set in C4.5) from weeks $N-1$ to $N-19$ (190 objects). 
Fig. \ref{comparativaC45} shows the percentage of correct predictions for our system and C4.5 classifier, using both attribute selections.
Other cols are also shown: 'user's most voted results', local team always win and two random generated. These 'random generated' cols were built assuming different weigths per result. It means, 
$<1:55\%,X:23\%,2:22\%>$ and $<1:65\%,X:18\%,2:17\%>$ were used, where $1,X,2$ are the probabilities for forecasting a match with the result: local team wins, drawn and away team wins, respectively.

It is worth to note that, while classifier achieves highest performances (58,68\%) when number of matches increase from 190 to 380, our system reaches this highest performance (59,74\%) using only 190 instances. 
This conclusion is based on our system use some {\em fast and frugal} \cite{prob} methods, and these
are designed to achieve aceptable results using as less as possible resources.

The relationship of our proposal with Recognition Heuristics \cite{ecological} (roughly speaking, if one of the  possibilities is recognized and the
other is not, then infer that the recognized object has the higher value
with respect to the criterion) is not clear. We may assert that our model \simon{recognize}{recognises} trends in contexts. Trends (represented as association rules) can be considered as a kind of recognizing method, though. {\ttg The system is based on bounded rationality models instead of statistic models, although in future hybrid models will be considered.}

{\ttg
In the short term, we carry on extending our system in order to be able to combine the results of two or more attribute sets with different exigency level. Therefore the system will return only one result and more reliable. In the long term, we aim to extend the model in orfer to obtain a general system to detect emergent concepts in Complex Systems}

%A corto plazo, nuestro trabajo se centra en extender el sistema para que la aplicación combine dos conjuntos de atributos distintos,
%uno mas estricto y otro menos, para una misma consulta
%La aplicación lanzaría primero el contexto estricto y si este no obtuviera ningún resultado entonces lanzaría el segundo.
%Currently the system allows a limited logical combination of attributes.

{\ttg
After some real betting experiments during current season (2010-2011) with one customized attribute set, we have observed
another intriguing fact. 
If we take a look to number of successful predictions per week, we are able to distinguish some groups of consecutive weeks in which number of correct predictions is under or over the average. 
Recall that these predictions are the logical inferred results by one customized attribute set. This suggests that it could be possible to find another attribute set, with a different parameters customization, which it will accomplish the correct predictions of first
attribute set. 
It means that when first attribute set produce bad forecasting, second should produce good ones, and vice versa.
The reason of this is that each match there is not only one possible logical result. It means, when
one of firsts teams of current ranking plays against one of lasts team, attending to ranking criteria, the logical result of this match would be that first one wins. But if we attend to others, like first team lost last week and second team won last 5 weeks, this results would be different. 
Future works pass through for finding these complementary attribute sets and detecting when their behaviors change during season in order to select the proper attribute set to forecast each week.

%DUDA: Seria interesante comentar tambien lo de que esto podria servir para sugerir apuestas dobles y triples, el problema es que
%eso es algo propio de las quinielas y aqui se habla de apuestas de futbol en general.

%NUEVO:Tras los experimentos realizados en la temporada actual es posible observar en algunas semanas consecutivas (no siempre),
%tendencias mas 'logicas' y tendencias menos 'logicas'. Esto es, hay rachas de 3 o 4 semanas consecutivas con numero de
%aciertos similares, no existen apenas picos
%individuales con muchos aciertos o con muchos fallos, van en grupo.Para ser exactos, estas tendencias son mas o menos logicas con
%respecto al conjunto de atributos usado. Por lo que seria interesante tener varios conjuntos de atributos que fueran mas o menos
%disjuntos en las predicciones que sugieren (siempre con cierta logica claro y no en todos los partidos pero si en bastantes). Hay
%que recordar que no siempre existe un solo resultado logico, un claro ejemplo es cuando se enfrenta un equipo de los primeros en la
%clasificacion con otro de los ultimos clasificados, la logica diria que gana el que esta bien clasificado, pero y si ademas sabemos
%que el primer equipo perdió el ultimo partido mientras que el equipo que esta por debajo en al clasificacion lleva 4 victorias
%consecutivas? La logica de la clasificacion diria que gana el primero y la logica de las vitorias consecutivas
%diria que gana el segudno. Ademas este tipo de situaciones representan una forma interesante para elegir dobles y triples.
}
{\ttg
%no tengo claro como empezar esta frase.
Finally, we are also analyzing how to finde a weight for matches which allows the system to work with matches from different divisions, simultaneously. Note that a winning match at first division will have a higher weight than a winning at second.
This will be really useful at the beginning of season because of we need to compute attributes related to previous matches results 
and teams which are involved played at different divisions last season.
}
%Es necesario analizar la posibilidad (y cálculo) de un factor de penalización para el caso de un equipo ascendido a
%una categoría superior y
%el sistema tomará los datos sobre la racha de partidos del equipo en las útlimas
%jornadas de la temporada anterior. Estamos experimentando para estimar dicho factor.

\section{Acknowledgements}

Supported by TIN2009-09492 project of Spanish
    Ministry of Science and Innovation, and {\em Excellence project} TIC-6064 of {\em Junta de Andalucía}  cofinanced with FEDER founds.

\footnotesize
\bibliographystyle{apalike}

%%%%%%%%%%%%%%%% ORIGINAL %%%%%%%%%%%%%%%%%%%%
\bibliographystyle{apalike}
%\bibliography{sport}

\begin{thebibliography}{99}

%\bibitem{paella} G.A. Aranda-Corral (Re)Organizing web search results by means of semantic and visual tools. %4th Workshop on the Future of Web Search: Semantic Search
%Ibiza - April 17-18, 2009 
%\bibitem%[Amir \&  McIlraith (2005)]
%{Amir} E. Amir, S. McIlraith,
%Partition-based logical reasoning for first-order and propositional
%theories, {\em Artificial Intelligence} {\bf 162}(1-2) 49-88 (2005).
%\bibitem{Hale} Why Spain will win..., Engineering \& Technology 5 June - 18 June 2010
\bibitem[Inst. Engineering and Technology 2010] {Hale} Why Spain will win..., Engineering \& Technology 5 June - 18 June 2010.
\bibitem[Alonso et al. 2008] {CLA} J. A. Alonso-Jiménez, G. A. Aranda-Corral, J. Borrego-Díaz, and M. M. Fernández-Lebrón, M. J. Hidalgo-Doblado, Extending Attribute Exploration by Means of Boolean Derivatives, Proc. 6th Int. Conf. Concept Lattices and Their Applications (CLA2008), pp. 121-132 (2008).
\bibitem[Andersson et al. 2003]{frugal1} P. Andersson, M. Ekman, J.  Edman,
Forecasting the fast and frugal way: A study of performance and information-processing strategies of experts and non-experts when predicting the World Cup 2002 in soccer,
Working Paper Series in Business Administration 2003:9, Stockholm School of Economics.
\bibitem[Aranda-Corral \& Borrego-Díaz 2010]{hais} G. A. Aranda-Corral, J. Borrego-Díaz, Reconciling
Knowledge in social tagging web services. Proc. 5th
Int. Conf. Hybrid AI Systems (HAIS 2010), LNAI, vol. 6077. Springer-Verlag, Berlin, 383-390 (2010).
%\bibitem{calculemus} G. A. Aranda-Corral, J. Borrego-Díaz, M. M. Fernández-Lebrón, Conservative Retractions of Propositional Logic Theories by Means of Boolean Derivatives: Theoretical Foundations. Proc. 16th CALCULEMUS Congress. LNAI, vol. 5625. Springer-Verlag, 45-58 (2009).
\bibitem[Aranda-Corral et al. 2011]{Torre} G. A. Aranda-Corral, J. Borrego-Díaz, J. Galán-Páez, Confidence-Based Reasoning with Local
Temporal Formal Contexts. to appear in IWANN 2011, LNCS (2011).
\bibitem[Aranda-Corral et al. 2011b]{dali} G. A. Aranda-Corral, J. Borrego-Díaz, J. Galán-Páez, Bounded Rationality for Data Reasoning based on Formal Concept Analysis. To appear in DEXA Workshop DALI (2011).
\bibitem[Armstrong 1974]{Armstrong} W. Armstrong, Dependency structures of data base relationships. Proc. of IFIP
Congress, Geneva, 580-583 (1974).
\bibitem[Balcázar 2010]{balcazar} J.L. Balcázar, Redundancy, Deduction Schemes, and Minimum-Size Bases for Association Rules, Logical Methods in Computer Science 6(2):1-23  (2010).
%\bibitem{survey} C. Bettini, O. Brdiczka, K. Henricksen, J. Indulska, D. Nicklas, A. Ranganathan, and D. Riboni, A survey of context modelling and reasoning techniques. Pervasive Mob. Comput. 6(2):161-180  (2010)
%\bibitem{conflicts} A. Bikakis,  G. Antoniou,  P. Hassapis,  Strategies for Contextual Reasoning with %Conflicts in Ambient Intelligence. To appear in Knowledge and Inf. Systems (2010).
%\bibitem{handbook} El del handbook
%\bibitem{ESWC} J. Borrego-D\'{\i}az and A. M. Ch\'{a}vez-Gonz\'{a}lez. Visual ontology cleaning: Cognitive 
%principles and applicability. In 3 European Semantic Web Congress, Lecture Notes in Computer Science n. %4011, pp. 317?331(2006). 
%\bibitem{Constantinou} A. Bikakis, and G. Antoniou, Distributed Defeasible Contextual Reasoning in Ambient Computing. In Proc. European Conf. on Ambient intelligence. Lecture Notes In Computer Science, vol. 5355. Springer-Verlag, Berlin, Heidelberg, 308-325 (2008)
%\bibitem{Buvac} S. Buvac and I. A. Mason. Propositional logic of context. Proc. AAAI 93, pages 412-419, 1993.
\bibitem[Brunswik 1955]{Bru} E. Brunswik, Representative design and probabilistic theory in
a functional psychology. Psychological Review, (62):193-217 (1955).
\bibitem[Ganter \& Wille 1999]{FCA} B. Ganter and R. Wille. Formal Concept Analysis - Mathematical Foundations. 
Springer, 1999.
%\bibitem{local} C. Ghidini, F. Giunchiglia, Local models semantics, or contextual reasoning = locality + compatibility. Artif. Intell. 127, 2 (Apr. 2001), 221-259.
\bibitem[Giarratano \& Riley 2005]{giar} J. C. Giarratano, G.D. Riley, Expert Systems: Principles and Programming. Brooks/Cole Publishing Co ( 2005). 
\bibitem[Goldstein \& Gigerenzer 1996]{prob} D. G. Goldstein, G. Gigerenzer, Reasoning The Fast and Frugal Way: Models of Bounded Rationality, Psychological Review  103(4): 650-669 (1996).
\bibitem[Goldstein \& Gigerenzer 2002]{ecological} D. G. Goldstein, G. Gigerenzer, Models of ecological rationality: the recognition heuristic, Psychological review, 109(1): 75-90 (2002).
\bibitem[Goldstein \& Gigerenzer 2009]{frugal} D.G. Goldstein, G. Gigerenzer,  Fast and frugal forecasting. International Journal of Forecasting, 25, 760-772 (2009).
\bibitem[Guigues \& Duquenne 1986]{stem} Guigues, J.-L., Duquenne, V.: Familles minimales d’ implications informatives resultant
d’un tableau de donnees binaires. Math. Sci. Humaines 95, 5–18 (1986).
%\bibitem[Hunter 2000]{hunter} A Hunter, Reasoning with inconsistency in structured text, Knowledge Engineering Review, 15(4):317-337 (2000)
\bibitem[Imberman et al. 1999]{bool} S. P. Imberman, B. Domanski, R. A. Orchard: Using Booleanized Data To Discover Better Relationships Between Metrics. Int. CMG Conference 1999: 530-539
%\bibitem{seminal} J. McCarthy, J. 1993. Notes on formalizing context. In Proceedings of the 13th international Joint Conference on Artifical intelligence. Morgan Kaufmann Publishers, San Francisco, CA, 555-560 (1993).
\bibitem[Min et al. 2008]{sd4} B. Min, J. Kim, C. Choe, H. Eom, R. I. McKay, A compound framework for sports results prediction: A football case study. Know.-Based Syst. 21(7):551-562. 2008
%\bibitem{temp} R. Neouchi, A.Y. Tawfik, R. A. Frost. 2001. Towards a Temporal Extension of Formal Concept Analysis. Proc.14th Conf. Canadian Soc. on Comp. Studies of Intell., Lecture Notes in Computer Science  2056, pp.  335-344 Springer-Verlag, 2001.
%\bibitem[Neouchi et al. 2001]{temp} R. Neouchi, A.Y. Tawfik, R. A. Frost. 2001. Towards a Temporal Extension of Formal Concept Analysis. Proc.14th Conf. Canadian Soc. on Comp. Studies of Intell., Lecture Notes in Computer Science  2056, pp.  335-344 Springer-Verlag, 2001.
%\bibitem[Pachur \& Biele 2007]{Pachur} Pachur, T., Biele, G. (2007). Forecasting from ignorance: The use
%and usefulness of recognition in lay predictions of sports events.
%Acta Psychologica, 125, 99–116.
%\bibitem[Simon 1982]{simon} H.A. Simon, Models of bounded rationality. Cambridge, MA: MIT Press (1982).
\end{thebibliography}
%%%%%%%%%%%%%%%% ORIGINAL %%%%%%%%%%%%%%%%%%%%

\end{document}